# Natural Scene Image Annotation Using Local Semantic Concepts and Spatial Bag of Visual Words

Yousef Alqasrawi

*Faculty of Information Technology, Applied Science Private University, Amman, Jordan*

**Abstract:** The use of bag of visual words (BOW) model for modelling images based on local invariant features computed at interest point locations has become a standard choice for many computer vision tasks. Visual vocabularies generated from image feature vectors are expected to produce visual words that are discriminative to improve the performance of image annotation systems. Most techniques that adopt the BOW model in annotating images declined favorable information that can be mined from image categories to build discriminative visual vocabularies. To this end, this paper introduces a detailed framework for automatically annotating natural scene images with local semantic labels from a predefined vocabulary. The framework is based on a hypothesis that assumes that, in natural scenes, intermediate semantic concepts are correlated with the local keypoints. Based on this hypothesis, image regions can be efficiently represented by BOW model and using a machine learning approach, such as SVM, to label image regions with semantic annotations. Another objective of this paper is to address the implications of generating visual vocabularies from image halves, instead of producing them from the whole image, on the performance of annotating image regions with semantic labels. All BOW-based approaches as well as baseline methods have been extensively evaluated on 6-categories dataset of natural scenes using the SVM and KNN classifiers. The reported results have shown the plausibility of using the BOW model to represent the semantic information of image regions and thus to automatically annotate image regions with labels.

## 1. INTRODUCTION

The availability of low-cost digital cameras and mobile devices and their attractive ability to upload and share photos to photo-sharing websites have led to a dramatic increase in the size of image collections. For efficient use of such large image collections, image annotation, searching, browsing and retrieval techniques are required [1-3]. Having images automatically annotated with labels, images can be retrieved in the same way as text documents.

Automatic image annotation is considered a promising approach to bridge the semantic gap between low-level image features and high-level semantics through extracting semantic features or concepts from images using machine learning techniques [4]. Image annotation helps to understand image visual content. In this paper, semantics are text words used in describing the visual content of an image.

In general, automatic image annotation systems learn the semantic concepts of a set of images by associating low-level features to high-level concepts. The trained system is then used to predict a set of semantic concepts to annotate previously unseen test images. Two major aspects are related with automatic image annotation: feature extraction and semantic concepts learning. Feature extraction can be performed at two levels: image level and region level. At image level, the visual content of an image is described by global features such as color histograms. However, global features do not describe different parts of an image. At region level, images are initially partitioned into rectangular blocks or regions via fixed size grid or using a segmentation algorithm. Each block or region is represented by features extracted from it, such as color and texture. Thus, an image with n blocks or regions is represented by a set of *N* feature vectors. Then, automatic image annotation system is trained, using a machine learning approach such as SVM, to assign each feature vector to a pre-defined category. Automatic image annotation is not an easy task as it is computationally expensive [4] and produce unrelated semantically similar regions [5]. For this reason, this work avoids image segmentation and adopt grid-based image division.

Recently, many approaches utilized local invariant features [6], local semantic concepts [7], and the bag of visual words (BOW) in many computer vision tasks and have shown improvements in the performance of scene image classification [6, 8, 9, 10-12]. There are two key phases to build an image classification system that use BOW technique. The first phase extract local features that characterize the visual content of the image. The work that is presented in this paper relates to this part. The second part is the classifier. The elements needed to build a bag of visual words are: feature detection, feature description, visual vocabulary construction and image representation, each step is performed independently of the others.

Recently, a range of new methods have been proposed to advance the performance of the traditional BOW model. It can be classified into three main types. The first type tries to improve the visual vocabulary construction [20-24]. The second type integrates multiple features with weighting techniques [11, 13-16]. In the third type, different techniques are proposed that incorporate spatial information with the BOW and have shown improvement in the performance of scene classification tasks [17-20].

Despite that most of these approaches achieved encouraging results in many computer vision tasks, no one claimed that his approach is the best. The complexity nature of natural scenes and different arrangements of items in image content means that, images with visually similar contents from two different classes are usually mis-categorised, (e.g., ambiguity in visual appearance between coasts and river/lake categories). We suggest that many of these problems could be better addressed by constructing a unified technique that utilizes knowledge from discriminative visual vocabularies, multiple image features and their spatial arrangements [21].

This paper proposes a framework for automatically annotating image regions with semantic keywords from a predefined vocabulary of words. The framework utilizes a hypothesis presented in [22]. The hypothesis claims that local semantic concepts in natural scenes correlates with local keypoints found in image regions. The proposed hypothesis investigates the correlation between the distribution of local semantic concepts and local keypoints detected in image regions and annotated with these semantic concepts. Based on this hypothesis, concept-based bag of visual words is proposed to represent the visual content of image regions. In this paper, concept-based bag of visual words is further improved by proposing to build visual vocabularies from image halves. Extensive experiments are conducted over a natural scene dataset with six categories.

## 2. RELATED WORK

Due to the semantic gap between image features and human interpretation of images, automatic image annotation comes as a suggested solution to bridge this gap. It works by extracting semantic keywords from images by using machine learning approaches [4].

In general, automatic image annotation systems consider learning the semantic concepts of a set of images by associating low-level features with high-level concepts. The trained system is then used to predict a set of semantic concepts to annotate previously unseen test images. There are many different techniques to carry out image annotation. According to [23], image annotation can be classified into three types:

- Classification-based annotation: in this type, image annotation is treated as classification problem using multiple classifiers, such as SVM, to learn image semantic concepts.
- Probabilistic-based annotation: in this type, probabilistic models are developed to estimate the relation between visual image contents and semantic concepts.
- Retrieval-based annotation: in this type, semantic concepts of images that are semantically relevant to the query image are employed to annotate the query image.

For automatic image annotation, higher level semantics can be learned from image sample or image regions represented with low level features. New images are labeled with semantic labels using the trained model. Automatic image annotation can be accomplished in three ways:

- Annotate images with single label.
- Annotate images with multi labels.
- Annotate images with metadata from the web.

In the first approach, image annotation is considered as a binary classification problem. Image features are first provided to a binary classifier, such as support vector machines, artificial neural network, or decision tree. The trained classifier is then used to label a new image with a semantic label. Support vector machines are the common choice for many classification problems, such as image classification, object recognition and text classification [24-27]. It has an advantage over other classifiers that it achieves optimal class boundaries by finding the maximum distance between the hyperplane of classes. The hyperplane is trained to separate the samples of one class from other class. For multi-class classification and annotation problem, two common schemes are used: one-vs.-one and one-vs.-all. Thus, multiple SVMs are used to learn each class individually such that a test image is labeled with a decision fused from decisions of all classifiers.

Chapelle et al. [25] proposed a framework to train 14 SVM classifiers to learn 14 image semantic concepts using one-vs.-all paradigm. Only color histogram is used to represent visual image content at three color channels of HSV color space. No image segmentation is employed in their work. Each SVM is trained on a particular semantic concept where all images belong to the same semantic concept are regarded as positive samples while the others are considered as negative. A new image is classified based on a voting approach to select the classifier with maximum probability value.

To improve the classification power of SVM, Goh et al. [26] proposed a three-level classification scheme using three different sets of SVMs, one-class, two-class, and multiclass, to annotate images with semantic classes. They developed a confidence-based approach to fuse the output of classifiers propagated at different levels based on the difference between the highest two output decisions. An image is assigned to the semantic class with highest cumulative confidence. Image annotation is performed at image level without any segmentation.

Cusano et al. [24] proposed a framework to annotate image regions with seven semantic concepts-sky, skin, vegetation, snow, water, ground, and buildings. Their image annotation framework uses multi-class SVM to assign/classify image regions into one of the seven semantic concepts. All images used in their experiments are divided into overlapped tiles around each pixel and are labeled with the seven semantic concepts. Image regions are described by color histogram in the HSV color space. The classifiers are trained on random sample of tiles chosen from each semantic concept. To annotate the whole image, they proposed a threshold-based strategy to accept or reject the semantic concept based on the decision obtained from SVM classifiers.

Multilevel SVM sets are also explored in [27] to annotate images with semantic concepts using both global features and local features extracted from the whole image and image regions, respectively. For region-based features, images are divided into 5 non-overlapping blocks. Both features are used in two different sets of SVMs. This approach also differs from the previous one in the way how predictions are fused to get the semantic label for the new image.

Single label image annotation approaches improve image retrieval by just typing the keywords related to the semantic concepts. So, there is no need to do image matching.

Multiple image annotation refers to approaches that assign more than one label to the whole image. To make images understandable by humans it is important to represent the semantic structure of images for semantic image annotation. The choice of the semantic keywords or concepts is dependent on the domain, user needs and the application. For multiple image annotation, this work focuses on using constrained vocabulary of a small size to locally annotate natural scene images with semantic concepts.

As has been mentioned earlier, single label image annotation is a classification problem. In contrast, multiple labels image annotation seeks to annotate image with multiple keywords, either to image regions or part of them. Barnard and Forsyth [28] proposed a generative hierarchical clustering model to assign multiple labels to an image through mapping labels to regions. Their model learns the joint statistics of words and regions and build word-region co-occurrence matrix to attach words to regions. Instead of using hierarchical clustering, Duygulu et al. [29] proposed a translation model to map regions to words. Both models used normalized cuts algorithm [30] to extract regions from images.

Aksoy et al. [31] addressed the relations between labeled regions. They proposed a visual grammar to represent the relation between image regions to reduce the semantic gap. Bayesian classifiers are used to label image regions then the grammar model is used to classify images into scene categories.

Mojsilovic et al. [32] addressed the problem of the semantic gap by introducing semantic indicators (sky, water, skin,…etc) based on experiments with human subjects. Global and local visual features are extracted from images and quantized into regions. These regions are names by human subjects. These semantic indicators are used then for scene categorization.

Recently, Makadia et al. [33] argue that most complicated state-of-the-art automatic image annotations techniques lack of comparisons with simple baseline measures to justify the need for such complicated models. They proposed a set of baseline methods, using color and texture, to automatically annotate images with keywords based on nearest neighbors. For color features, color histograms are generated from the three components of images represented in RGB, HSV and LAB color spaces, respectively. Gabor filters and Haar wavelets are used to represent image textures. To calculate nearest neighbors, three distance measure are jointly combined and evaluated using KL-divergence, L1-distance, and L2-distance. For test images, predicted labels are assigned to whole image without specifying to which region the labels refer to. Many surveys on automatic image annotation are available in literature and most recently are carried out by [4, 34].

## 3. NATURAL SCENE DATASET

This paper studies the task of automatically annotating image regions with local semantic keywords or concepts. To accomplish this task, we use a natural scene dataset of 700 images categorized into six scene categories [5]. The six categories are river/lakes, coasts, plains, forests, mountains, and sky/clouds. Authors of this dataset manually annotated image regions with one of the nine semantic concepts: sky, water, grass, trunks, foliage, field, rocks, flowers and sand. For validation and experimental work, images were partitioned into a matrix of (10×10) regions which produce 70000 regions (see Figure 1). In the original work of Vogel and Schiele [5], image regions with multiple semantic concepts were excluded in training and testing local semantic classifiers. Images in the dataset are organized into two folders. The first folder contains natural scene images. The second folder contains text files, each file contains semantic concepts of the corresponding natural scene images in the first folder. Assume that I={$i_1,i_2,…,i_N$} be the set of images from the first folder and T={$t_1,t_2,…,t_N$} is the set of text files, where N=700. A text file $t_j$ contains a list of annotations for an image $i_j$, where j=1,2,…,N. The distribution of local semantic concepts in each scene category is depicted in Table 1.

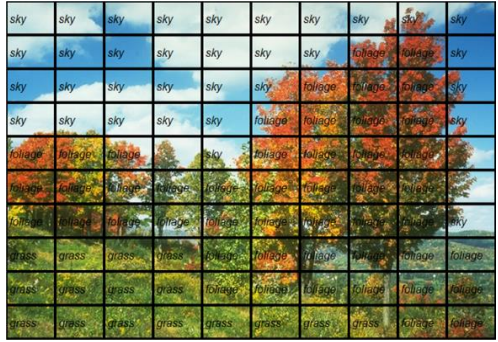

Figure 1: A 10x10 annotated image from the 'forest' scene category.

Table 1: Sizes of the nine local concept classes located in each scene category.

| | Sky | Water | Grass | Trunks | Foliage | Field | Rocks | Flowers | Sand |
|---|---|---|---|---|---|---|---|---|---|
| Coasts | 2960 | 4326 | 430 | 32 | 1284 | 194 | 1922 | 46 | 825 |
| Rivers/lakes | 1728 | 2826 | 273 | 82 | 2629 | 204 | 1553 | 12 | 0 |
| Forests | 335 | 39 | 465 | 1419 | 6464 | 309 | 47 | 31 | 0 |
| Plains | 2879 | 14 | 1608 | 36 | 964 | 2649 | 330 | 1898 | 897 |
| Mountains | 4416 | 54 | 649 | 56 | 2335 | 735 | 7401 | 62 | 23 |
| Sky/clouds | 2978 | 34 | 78 | 0 | 33 | 97 | 57 | 0 | 0 |
| # of image regions | 15296 | 7293 | 3503 | 1625 | 13709 | 4188 | 11310 | 2049 | 1745 |
| OVERALL | | | | | 60718 | | | | |

## 4. LOCAL SEMANTIC CONCEPTS VS. LOCAL KEYPOINTS

This section reviews our hypothesis proposed in [22] which investigates the correlation between local semantic concepts and local keypoints, based on their distributions over all image regions. Local semantic concepts are labels assigned to image regions. Using (10×10) fixed grid layout, an image is divided into 100 image regions. Low-level features, such as color and texture, are normally used to represent the visual content of image regions. These features are not invariant to different changes of the visual contents. Thus, to provide invariance to changes in illumination, rotation, etc, interest points can be used [35]. Interest points or local keypoints correspond to image structures that are considered important.

Using sparse representation approach, a keypoint detector, such as DoG detector [6], locates local keypoints that contain distinctive information in their surrounding area and should be invariant to geometric transformations. These keypoints are then described using robust and informative features such as SIFT.

Having the ground truth semantic labels provided with the natural scene dataset, and the local keypoints provided by an interest point detector, two facts can be drawn:

**1.** Each region in an image can be described by two corner locations; the top left corner $(x_1,y_1)$ and the bottom right corner $(x_2,y_2)$. The $(x_2-x_1)\times(y_2-y_1)$ is the size of the given region.

**2.** Applying Keypoint detector to an image would generates a list of coordinates of all the keypoints detected in the image. The detector also generates more features that describes the area around each detected keypoint, such as orientation and scale. We are interested on the coordinates of keypoints.

Now, we have data about the coordinates of all image regions and coordinates of all detected keypoints. The assumption is; the distribution of semantic keywords, used to annotate regions, may correlates with the distribution of detected keypoints in a particular scene category. This hypothesis can be justified as: for each natural scene category and semantic concept, we count the number of keypoints with coordinates located in the areas of all image regions labelled with the same semantic concept. If the distribution of semantic concepts is similar or close to the distribution of local keypoints then there is a possible correlation between them.

Figure 2 shows the distribution of local keypoints located in image regions of each semantic concept and over all scene categories.

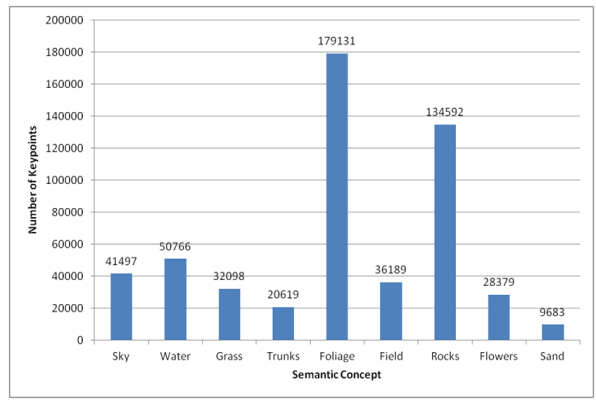

Figure 2: Distribution of local keypoints detected in image regions of each semantic concept over all scene categories.

The correlation between semantic concepts and local keypoints is shown in Figure 3. Also, the distributions of local semantic concepts and local keypoints over each of the six natural scenes are depicted in Figure 4 and Figure 5. The distribution of local semantic concepts at the upper and lower halves of all images is shown in Figure 6 whereas the distribution of local keypoints in image regions labelled with the semantic concepts is shown in Figure 7.

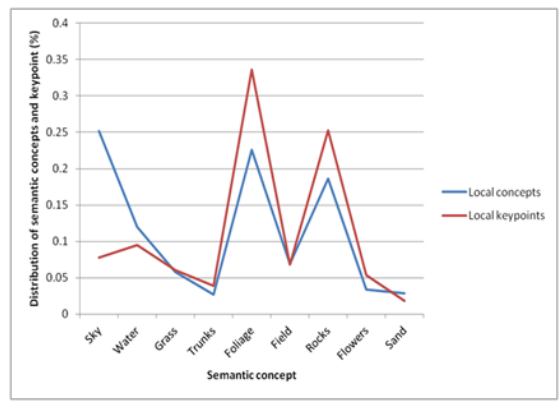

Figure 3: The correlation between the distribution (%) of local semantic concepts and local keypoints.

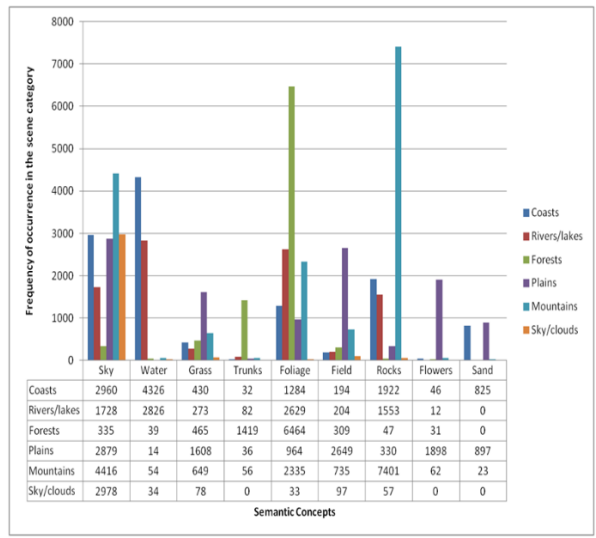

Figure 4: Distribution of each semantic concept over each category.

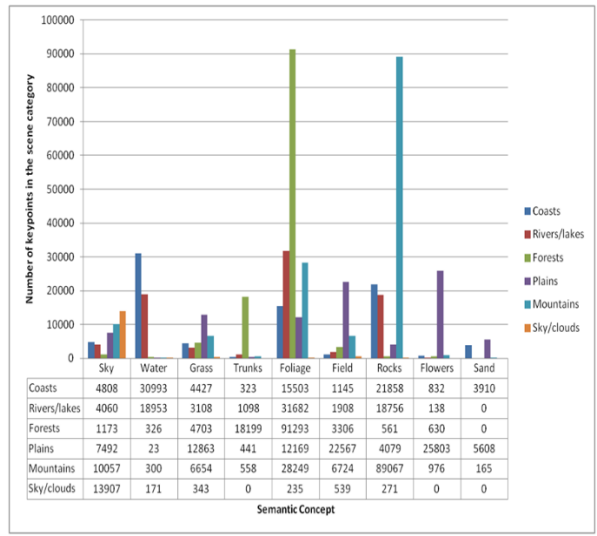

Figure 5: Distribution of keypoints located in regions of each semantic concept and over each scene category.

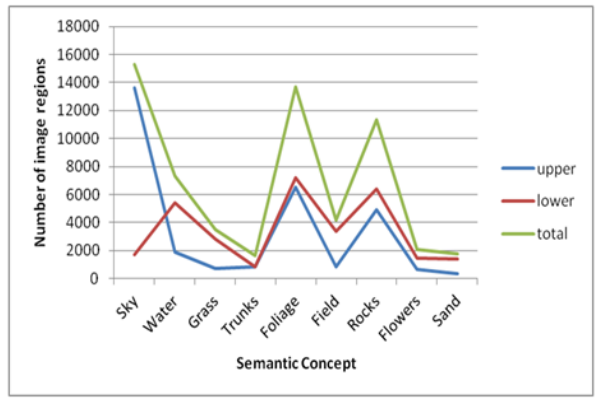

Figure 6: Distribution of image regions located in the upper and lower halves of images.

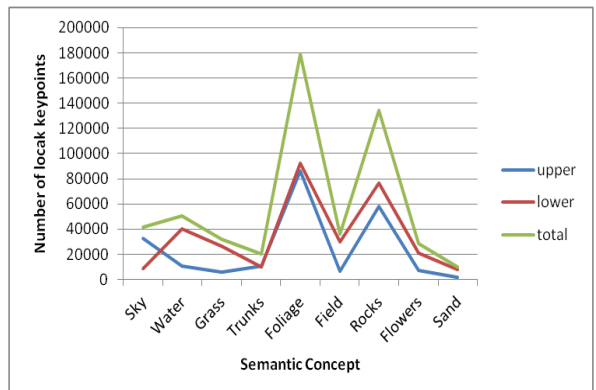

Figure 7: Distribution of local keypoints found in image regions in the upper and lower halves of images.

## 5. IMAGE ANNOTATION FRAMEWORK

In this section, a framework for annotating image regions with local semantic concepts based on bag of visual words is presented. The framework consists of two parts as depicted in Figure 8.

In the first part, DoG detector and SIFT descriptor are employed to find and represent local keypoints in all images of the dataset. Each SIFT descriptor is a vector of size 128-D. For each natural scene category, k-means algorithm, based on the Euclidean distance metric, is applied to all descriptors to build visual vocabularies. These vocabularies are then aggregated to form an integrated visual vocabulary of size (K×M) where K is the vocabulary size and M is the number of scene categories. We followed the same approach mentioned in [21] in building visual vocabularies for the six scene categories dataset mentioned in Section 3.

In the second part, image regions are represented using BOW, i.e., a histogram is generated for each region where each bin corresponds to the frequency of occurrence of each visual word in that image region. In this paper, bag of visual words (BOW) generated from an image region is called Concept-based Bag of Visual Words (CBOW). After representing all image regions using concept-based bag of visual words, and noting that image regions are labeled with semantic concepts, the CBOW histograms are used to learn classifiers to label image regions located in the test dataset with local semantic concepts.

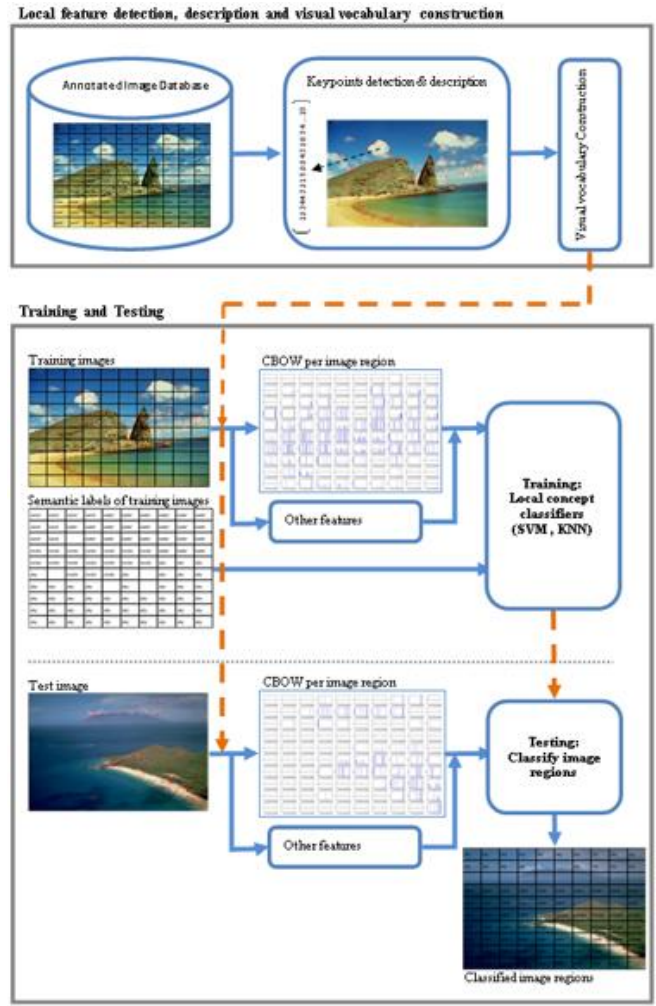

Figure 8: Flow diagram of the proposed image annotation framework.

### 5.1. Image visual vocabulary construction

The visual vocabulary on which CBOW rely on is of great importance. Building visual vocabularies from each scene category have shown improvements in the performance of natural scene classification task. To apply the same approach in building visual vocabularies to local semantic concepts, the number of visual vocabularies becomes larger. This is because of the fact that the number of local semantic concepts are usually larger than the number of scene categories these semantic concepts belong to. Therefore, it is worth to explore the influence of applying visual vocabularies obtained from natural scene categories (globally obtained from natural categories) to map SIFT features located in image regions (locally used to represent regions) to the indices of visual words and thus build CBOWs histograms.

Another choice of improving the power of visual vocabulary is to incorporate spatial information about local keypoints when clustering their features. Natural scene images, such as *coasts*, contain semantic concepts that usually appear in common places. For example, the *sand* concept can be found at the bottom of an image whereas the *sky* and *water* concepts appear at the top. Also, building visual vocabularies using k-means algorithm may lead to group dissimilar local keypoints in the same cluster. To avoid this problem, images are partitioned into two halves of equal size: upper and lower. An image of size W×H is divided into two halves, each has a dimension of (W/2)×(H/2). In this case, two integrated visual vocabularies are generated: Upper integrated visual vocabulary and Lower integrated visual vocabulary.

Upper integrated visual vocabulary is generated by clustering SIFT features located at the upper half of all training images. Lower integrated visual vocabulary is generated by clustering SIFT features located at the lower half of all training images. This approach has two main advantages:

**1.** Create more representative visual vocabularies about natural scenes that benefits from the spatial information of the keypoints and at the same time reduce ambiguities between clusters resulted from the clustering step.

**2.** Reduces the clustering time while maintaining or improving the discriminative power of the clusters. Preliminary experimental work has shown that quantizing local features at different spatial levels reduce the clustering time as well as improving the discriminative power of visual words or clusters. But, it increases the dimensionality of the BOW.

Comparing the annotation performance of the proposed framework using universal visual vocabulary, integrated visual vocabulary and upper and lower visual vocabularies allows us to analyse indirectly the dependence of bag of visual words representation on spatial locations of local keypoints. Also, to study the influence of using global visual vocabularies to represent image regions.

### 5.2. Image region representation

Two types of features are used to represent the visual content of image regions. In the first type, image regions are represented using concept-based bag of visual words (CBOW). These CBOW are generated using different visual vocabularies, as described in the previous section. In the second type, multiple low-level features are used as baseline methods and also to expand the discriminative power of CBOW.

### 5.3. Concept-based bag of visual words (CBOW)

In this paper, bag of visual words BOW generated at region level are called concept-based bag of visual words (CBOW). Each image in the dataset is spliced into 10x10 sub-regions.

Having images portioned into 10x10 regions, the next step is to build CBOW histogram for each image region. To do that, first, local keypoints are automatically detected in the whole image. Second, local keypoint features are extracted from each region which is defined around those keypoints. Third, construct visual vocabularies from detected descriptors. Fourth, map local keypoint to the index of the most similar visual word and their frequencies are counted to construct the CBOW for each image region *r*.

There are four kinds of visual vocabularies constructed and used in this paper to build the CBOW from image regions. We aim to compare and analyze their performances to annotate image regions. This is mentioned in the experimental work section. These visual vocabularies are:

- Universal visual vocabulary: To obtain the visual vocabulary, we use feature vectors (SIFT features) stored in image features database. All feature vectors from all training images on the dataset are quantized, using the k-means algorithm, to obtain centroids or clusters. These centroids represent visual words. Theses visual words constitute the universal visual vocabulary. This visual vocabulary contains 200 visual words.

- Integrated visual vocabulary: For the integrated visual vocabulary, SIFT features from all training images of each scene class are clustered into visual words. This visual vocabulary contains 1200 visual words (number of scene classes × 200 clusters).

- Universal Upper and Lower visual vocabulary: the upper visual vocabulary contains 200 visual words and the lower visual vocabulary contains 200 visual words.

- Integrated Upper and Lower visual vocabulary: the upper visual vocabulary contains 1200 visual words and the lower visual vocabulary contains 1200 visual words.

Let $I = \{i_1, i_2, ..., i_N\}$ be the set of all images in the dataset and that each image $i_k$ is divided into 10x10 sub-images such that $i_k = \{i_{k_1}, i_{k_2}, ..., i_{k_{100}}\}$, where $i_{k_r}$ is the r-th region in image $i_k$. In Algorithm 1, universal visual vocabulary generated for the natural scene dataset with 6 scene categories is used to build CBOWs from image regions. This vocabulary has been used in [21] for image classification. This vocabulary contains 200 visual words generated by clustering all SIFT features extracted from training images.

Algorithm 1: An algorithm to build CBOW from images regions using universal visual vocabulary.

**Input:** Use the *universal visual vocabulary* $V$ to construct CBOW from local image regions as follows:

**For** all images $I = \{i_1, i_2, ..., i_N\}$ in the dataset **Do**:

  a. Apply DoG feature detection technique to locate interest points in image $i_k$.
  b. Extract SIFT features (128-D) from each located keypoint.
  c. **For** each image region $i_{k_r}$ in the image $i_k$ **Do**:

      Quantize all SIFT descriptors located in region $i_{k_r}$ into one of the $M$ visual words, where $M$ is the size of the visual vocabulary $V$. An image region $i_{k_r}$ is then represented as a histogram $h_{k_r}$ of the frequencies of visual words.
    end

end

**Output**: a collection of $h_{k_r}$ histograms, where each $h_{k_r}$ is the CBOW for image $k$ at region $r$.

In Algorithm 2, the integrated visual vocabulary generated by clustering all SIFT features of training images over each scene category is used to build CBOW. This vocabulary has been used in [21] for the same dataset. It is important to study the influence of building visual vocabularies from local features located in parts of an image rather the whole image and compare their performances with the traditional visual vocabularies.

Algorithm 2: An algorithm to build CBOWs from images regions using integrated visual vocabulary.

**Input:** Use the *integrated visual vocabulary* $V$ to construct CBOWs from local image regions as follows:

**For** all images $I = \{i_1, i_2, ..., i_N\}$ in the dataset **Do**:

  a. Apply DoG feature detection technique to locate interest points in image $i_k$.
  b. Extract SIFT features (128-D) from each located keypoint.
  c. **For** each image region $i_{k_r}$ in the image $i_k$ **Do**:

      Quantize all SIFT descriptors located in region $i_{k_r}$ into one of the $M$ visual words, where $M$ is the size of the visual vocabulary $V$. An image region $i_{k_r}$ is then represented as a histogram $h_{k_r}$ of the frequencies of visual words.
    end

end

**Output**: a collection of $h_{k_r}$ histograms, where each $h_{k_r}$ is the CBOW for image $k$ at region $r$.

As mentioned earlier, clustering local descriptors located at the upper halves of images may generate a better quality clusters, i.e., their members are more semantically similar. It aims to reduce inter-class similarity and increase intra-class similarity throughout clustering features at image parts.

Although images can be divided to any number of tiles, only two parts from images are considered in this work; the upper half and the lower half. In other words, 50% of the image content, the upper part, is used to build the upper visual vocabulary whereas the other 50%, the lower part, is used to build the lower visual vocabulary. Upper and lower visual vocabularies are generated at two levels. The first level considers building upper and lower visual vocabularies from SIFT descriptors of all scene categories whereas in the second level visual vocabularies are generated from SIFT descriptors of each scene category, as shown in Figure 9.

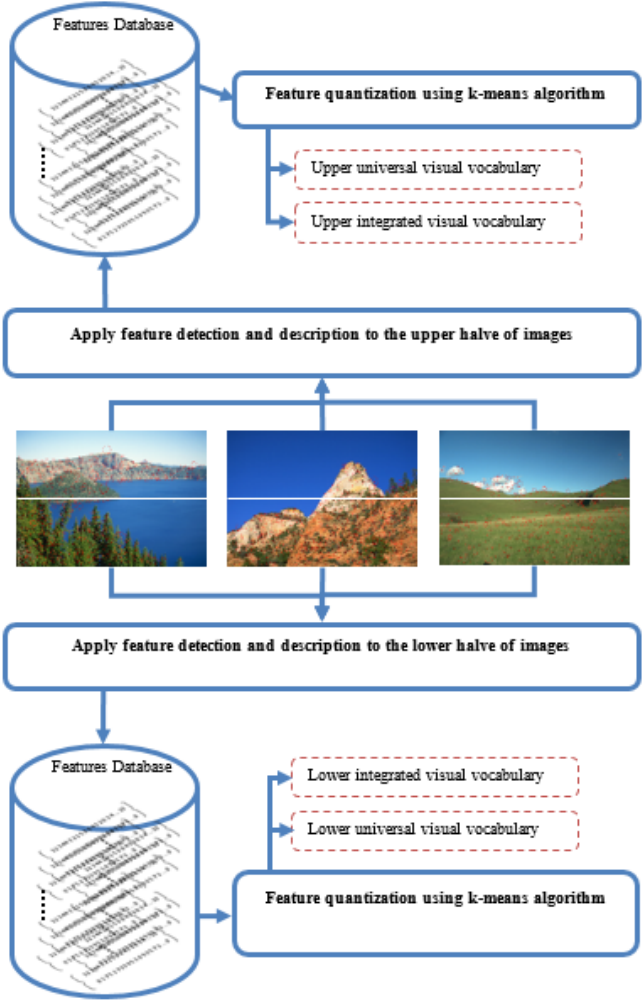

Figure 9: Upper and lower visual vocabularies construction. Images in the middle are samples from all training images used in the construction process.

Algorithm 3: An algorithm to build CBOWs from images regions, at upper halve, using universal visual vocabulary.

**Input:** Use the *upper universal visual vocabulary V*, to construct CBOWs from local image regions as follows:

**For** all images $I = \{i_1, i_2, ..., i_N\}$ in the dataset **Do**:

    a. Apply DoG feature detection technique to locate interest points in image $i_k$.
    b. Extract SIFT features (128-D) from each located keypoint.
    c. **For** each image region $i_{k_r}$ in the upper halve of image $i_k$, where *r=1,2,...,50* **Do**:

        Quantize all SIFT descriptors located in region $i_{k_r}$ into one of the $M$ visual words, where $M$ is the size of the visual vocabulary $V$. An image region $i_{k_r}$ is then represented as a histogram $h_{k_r}$ of the frequencies of visual words.
    **end**

**end**

**Output:** a collection of $h_{k_r}$ histograms, where each $h_{k_r}$ is the CBOW for image *k* at region *r*.

Having all visual vocabularies generated from upper and lower halves of images, the next step is to use them to build CBOW from image regions. Algorithms 3 and 4 illustrate the required steps to build CBOW from images regions at upper halves of images.

Algorithm 4: An algorithm to build CBOWs from images regions, at upper halve, using integrated visual vocabulary.

**Input:** Use the *upper integrated visual vocabulary V*, to construct CBOWs from local image regions as follows:

**For** all images $I = \{i_1, i_2, ..., i_N\}$ in the dataset **Do**:

    a. Apply DoG feature detection technique to locate interest points in image $i_k$.
    b. Extract SIFT features (128-D) from each located keypoint.
    c. **For** each image region $i_{k_r}$ in the upper halve of image $i_k$, where *r=1,2,...,50* **Do**:

        Quantize all SIFT descriptors located in region $i_{k_r}$ into one of the $M$ visual words, where $M$ is the size of the visual vocabulary $V$. An image region $i_{k_r}$ is then represented as a histogram $h_{k_r}$ of the frequencies of visual words.
    **end**

**end**

**Output:** a collection of $h_{k_r}$ histograms, where each $h_{k_r}$ is the CBOW for image *k* at region *r*.

To avoid repeating the same algorithms for the lower halves, lower visual vocabularies and image regions at the lower halves (r=51, 51,..., 100) are replaced with the corresponding ones used in Algorithms 3 and 4.

**5.4. Local from Global CBOW**

Constructing CBOW from visual vocabularies generated from each scene category allows us to study the relationship between natural scene categories and local semantic concepts. The principle of building integrated visual vocabulary is to quantize local features from each scene category. Thus, in the case of local semantic concepts, integrated visual vocabularies should be generated from local features located in image regions labeled with these semantic concepts.

The main aim of this section is to analyze the relationship between visual vocabularies, generated from scene categories, and CBOWs histograms generated from local image regions. The analysis is based on the distribution of all CBOWs histograms generated from image regions labeled with each of the semantic concepts. Also, we show that integrated visual vocabularies are more suitable to represent local features in image regions rather than the universal visual vocabulary. Furthermore, the distributions of CBOWs generated from upper and lower integrated visual vocabularies are analyzed.

To analyze the relationship between a visual vocabulary and the semantic concepts, all CBOW histograms generated from a visual vocabulary are summed up. Given that all image regions are annotated with semantic concepts and that each image region is represented by a CBOW then it is possible to sum up all CBOWs histograms for each semantic concept. For example, the distribution of all CBOW generated from image regions labeled with the semantic concept *sky*, using universal visual vocabulary, is shown in Figure 10. The same figure shows the distributions of all CBOW for *water*, *grass*, *trunks*, *foliage*, *field*, *rocks*, *flowers*, and *sand*. These distributions do not show any relation of the natural scene categories with the local semantic concepts.

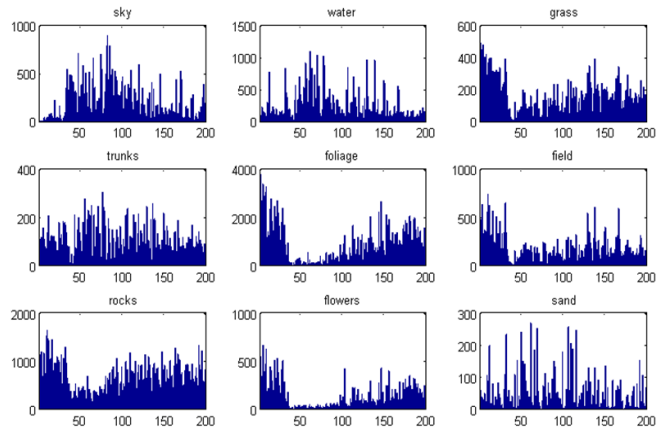

Figure 10: Sum of the CBOW histograms obtained using universal visual vocabulary at image level.

In contrast, Figure 11 shows CBOW distributions of image regions using integrated visual vocabulary. The integrated visual vocabulary contains 1200 visual words in which each 200 visual words represent 200 clusters generated from clustering all local descriptors located in images of specific scene category.

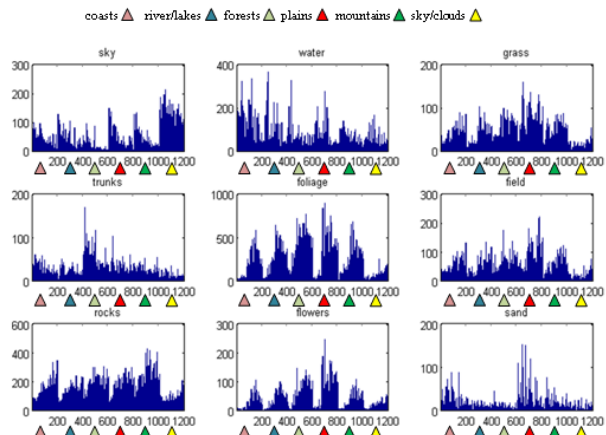

Figure 11: Sum of the CBOW histograms obtained using integrated visual vocabulary at image level.

The first 200 visual words represent clusters for the natural scene coasts, and the other 1000 visual words are clusters for the natural scene categories *river/lakes*, *forests*, *plains*, *mountains* and *sky/clouds*, respectively. This figure depicts interesting relationship between local semantic concepts and the natural scene categories. For example, large numbers of the local keypoints, found in image regions labeled with the semantic concept *sky*, are assigned to the last 200 visual words of the CBOW histograms. The last 200 visual words are clusters belonging to the scene category *sky/clouds*. And it is usual to see *sky* areas in *sky/clouds* natural scenes. Another interesting indication in the figure is the plot for the semantic concept *water*. For this concept, large numbers of the keypoints are assigned to the first 200 visual words which actually belong to the natural scene *coasts*, and this is natural to have water in natural scene coasts. The concept *water* is also available in the *river/lakes* scene category depicted in the same plot. The same figure contains the distributions for the semantic concept *sand*. Many of the keypoints are located in the first 200 visual words and visual words (600-799).

To this end, there is a relationship between natural scene categories and the semantic concepts they contain which are analyzed using the distributions of CBOW obtained using the integrated visual vocabulary and as has been discussed above. Moreover, these relationships confirm the hypothesis, presented earlier, in a way that image regions labeled with a semantic concept contain keypoints that are more likely to appear in the distributions of the CBOWs histograms for that semantic concept. Furthermore, Figure 11 shows the plausibility of using visual vocabularies, generated from local descriptors at image level, to build CBOWs. This will be confirmed in the experimental results section.

In the case of building visual vocabularies from local keypoints descriptors located at the upper and lower halves of images, the same approach is followed to analyze their relationship with the local semantic concepts. For the upper halves of images, two visual vocabularies are generated: universal and integrated visual vocabularies from which CBOW histograms are generated. For the lower halves of images, two visual vocabularies are also constructed: universal and integrated visual vocabularies.

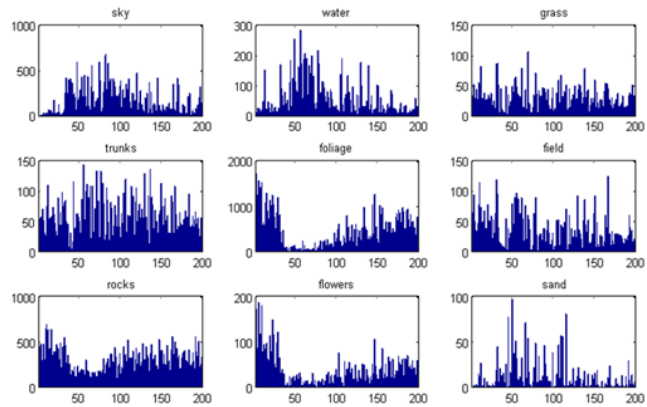

Figure 12: Sum of the CBOW histograms obtained using universal visual vocabulary at the *upper* half of the images.

Similar to using universal visual vocabulary in Figure 10, Figure 12 and Figure 13 show that the distributions of the CBOW over the nine semantic concepts do not show any indication from the distributions of the local semantic concepts on the upper and lower halves of images. Therefore no relationship can be inferred from both halves, though for image annotation, the performance of upper and lower CBOWs may still outperform CBOW at whole image level. This will be discussed in the experimental work section.

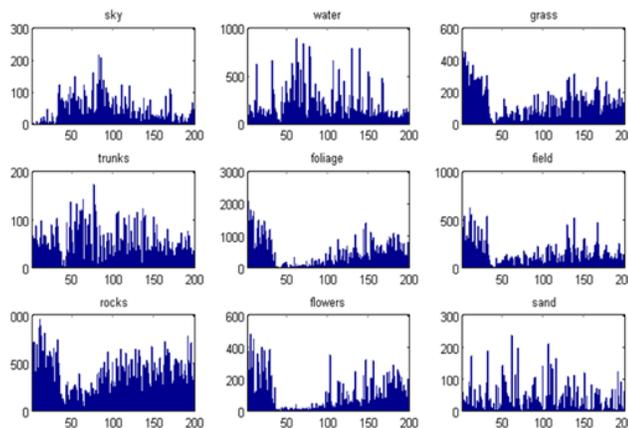

Figure 13: Sum of the CBOW histograms obtained using universal visual vocabulary at the *Lower* half of the images.

To reduce the ambiguity between visual words and improve their discriminative power, integrated visual vocabularies are generated from image halves. This is confirmed by analyzing the distributions of local keypoints in the CBOW over the nine semantic concepts and at both image halves. In Figure 14, the semantic concept grass is usually appear in the lower halves of images but could also appear at the upper half of the image, such as images of plains and mountains scenes. This is shown in the *grass* plot, where many of the local keypoints are indexed to the visual words (600-799) and (800-1000). These visual words represent clusters of *plains* and *mountains* scene categories.

For the *water* semantic concept, many of the keypoints are assigned to the visual words of the natural scene *coasts*. But, for the same semantic concept, water appears more in the lower halves of coasts and *river/lakes* scenes categories, as shown in Figure 15.

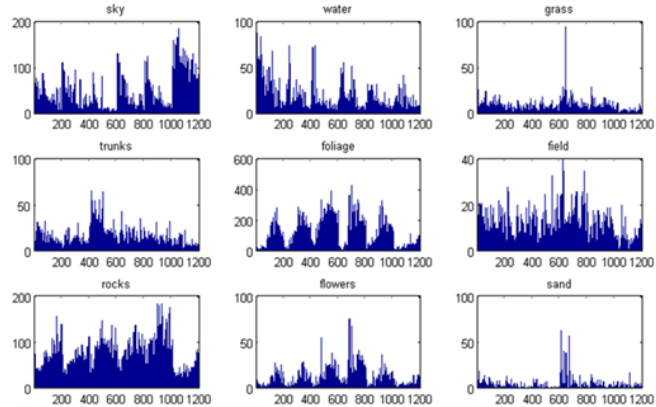

Figure 14: Sum of the CBOW histograms obtained using integrated visual vocabulary at the *Upper* half of the images.

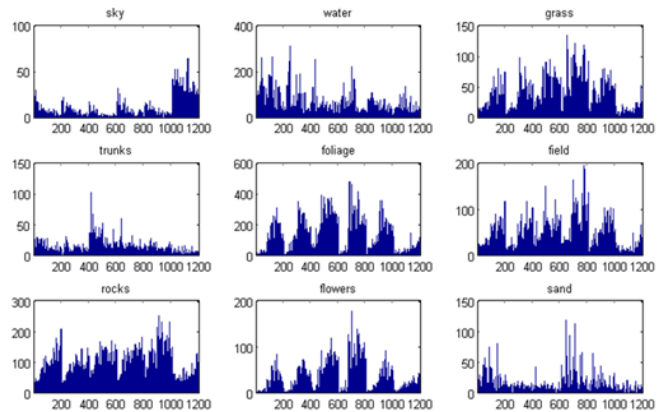

Figure 15: Sum of the CBOW histograms obtained using integrated visual vocabulary at the Lower half of the images.

It is interesting to use Figure 14 and Figure 15 to analyze the differences between the distributions of the CBOW at upper and lower halves. For example, local keypoints located in image regions of the *sand* semantic concept appear in the lower halve more than the upper halve which gives an indication of the importance of building multi-level visual vocabularies from image halves.

**5.5. Multiple features**

Motivated by the importance of using color information and textural featuers to describe the visual contents of natural scene images, this section presents a number of features that are used to improve the performance of CBOW image region representation and therefore will improve the performance of natural scene annotation task. Beside the CBOW, three types of features are chosen in this work. The first two features are devoted to represent color whereas the third feature is devoted to represent texture. Color histogram and color moments are used to extract color information from image regions. Discrete Wavelet Transform (DWT) is used to extract the textural features from image regions. To represent the visual contents of image regions, color moments, color histogram and DWT are used. Different combinations of these features are used without any weighting approach. Features are directly propagated to form a single feature vector.

In this paper, the wavelet decomposition is performed at image region level using 2 levels of decomposition using two-dimensional DWT functions available in MATLAB. Since images are converted into HSV color space, each image thus has three components: H, S and V. So, DWT is applied to each component of an image region separately. A feature vector of length 18 (3components × 6 energy measure) is therefore constructed from the three sub-bands (LH, HL, HH) at all resolutions and for each image region.

### 5.6. Prototypical local semantic concept representation

The idea of prototypical semantic concept representation is to associate local semantic concepts with instances that are considered to be prototypical with regards to their visual information [36]. Based on this idea, prototypical local semantic concepts are learned from visual information extracted from image regions. For the natural scene dataset used in this paper, there are nine semantic concepts. For each semantic concept, visual features extracted from image regions labelled with this semantic concept are averaged to obtain a local semantic prototype that represents instances of this semantic concept, hence this prototype does not necessarily to be member of the corresponding semantic concept. It is only a summary representation of each semantic concept. This result in generating nine local semantic prototypes learned from instances (visual features) of the nine semantic concepts. These prototypes will be used in the next section to annotate image regions using KNN approach.

## 6. EXPERIMENTAL SETUP

The first part of this section presents local semantic concept annotators, to annotate image regions, using KNN and SVM classifiers. It also presents the protocol that is followed to conduct the experiments. Experimental results are then reported with some discussion. The performance of local image annotators is assessed using the average precision produced from the confusion matrix.

### 6.1. Local Semantic Annotators

Image annotation at region level can be considered as a supervised classification problem. Each image region needs to be classified into one of predefined classes. In this paper, two classifiers are employed for the annotation task: the nearest neighbour approach (KNN) and support vector machines (SVM). The protocol used for all experimental work is as follows: Using the KNN classifier, all the experiments have been validated using 10-folds cross validation where 90% of all image regions are selected randomly for generating the local semantic prototypes and the remaining 10% are used for testing. Using SVM classifier, the same approach has been used where the 90% of all image regions used to generate the semantic concept prototypes are also used to train the SVM classifier whereas the 10% of image regions used in KNN are used for testing. The procedure is repeated 10 times for both classifiers such that an image region appears once in the testing part over the ten folds. This paper use the well-known LIBSVM [37] tool to implement the SVM classifier. The 10-fold cross validation technique is performed in each training set to choose SVM parameters. The Histogram intersection kernel is used to learn the SVM.

For the KNN approach, it is a pre-requisite to decide the similarity metric and the number of neighbors (K) that need to be visited to decide to which class an input instance belongs to, using a voting technique. Using the nine local semantic prototypes presented in Section 5.6, the KNN classifier needs to assign an input feature vector, which represents the visual content of input image region, to one of the local semantic concepts, thus K=1.

Algorithm 5: A step-by-step algorithm for generating semantic concept prototypes and how they are used to annotate image regions with local semantic concepts using KNN approach.

**Input:** (1) A collection of BOW histograms representing all image regions in the image dataset.

(2) A confusion matrix **mat** of size $N \times N$, where $N$ is the number of semantic concepts. Initialize **mat** to zeros.

**For** $i$=1 **to** 10 **Do**:

  a. Randomly select 90% of all BOWs histograms for training and 10% for testing.
  b. For each semantic concept $\bar{c}_j$, $j$=1,2,…, $N$, generate local semantic concept prototype $P_j$ by averaging all BOWs from the training set.
  c. Use KNN classifier, using the Euclidean distance and K=1, to find similarities between the BOWs in the test set and the local semantic prototypes $P$. An image region, represented by a BOW histogram, is assigned to the semantic concept $j$ such that the semantic prototype $P_j$ is the most similar prototype to the BOW of this image region.
  d. Compare the labels assigned to image regions from the test set with the ground truth labels. Report the results in the confusion matrix **mat.**

**end**

**Output**: Confusion matrix **mat**.

For simplicity, in this section the bag of visual words BOW and IBOW histograms are used to refer to the concept-based bag of visual words CBOW generated using the universal visual vocabulary and the integrated visual vocabulary, respectively. In other words, instead of using CBOW and CIBOW, the authors simply refers to them by BOW and IBOW. This section explicitly refers to the BOWs and IBOWs generated from image regions at the upper and lower halves in the experimental results. Algorithm 5 presents the work flow of annotating image regions with local semantic concepts using KNN and local semantic prototypes. The same algorithm is used to conduct different experiments using the local semantic prototypes approach.

Algorithm 6 presents the work flow of annotating image regions with local semantic concepts using KNN and local semantic prototypes generated from the upper and lower halves of images.

## 6.2. Experimental results

In this paper, four sets of experiments are conducted. The first two sets consider using universal and integrated visual vocabularies, obtained in [21], to build BOW and IBOW histograms from image regions. As mentioned at the beginning of the previous section, BOW and IBOW refer to concept-based BOW generated using both visual vocabularies. The last two sets of experiments investigate generating visual vocabularies from image halves to build BOW and IBOW for image regions at each halve separately. For fair comparisons, different types of features are included in the experiments and their annotation performances are reported.

Algorithm 6: A step-by-step algorithm for generating semantic concept prototypes at the upper and lower halves of images and how they are used to annotate image regions with local semantic concepts using KNN approach.

**Input:** (1) A collection of BOW histograms representing all image regions at the *upper* halve of images in the dataset.

(2) A collection of BOW histograms representing all image regions at the *lower* halve of images in the dataset.

(3) Two confusion matrices **mat_upper** and **mat_lower**, both of size $N \times N$, where $N$ is the number of semantic concepts. Initialize **mat_upper** and **mat_lower** to zeros.

**For $i$=1 to 10 Do:**

a. Randomly select 90% of all BOWs histograms, generated from the upper halves, for training and 10% for testing.
b. For each semantic concept $\bar{c}_j$, $j$=1,2,…, $N$, generate local semantic concept prototype $P_j\_upper$ by averaging all BOWs from the training set.
c. Use KNN classifier, using the Euclidean distance and K=1, to find similarities between the BOWs in the test set and the local semantic prototypes $P\_upper$. An image region at the upper halve, represented by a BOW histogram, is assigned to the semantic concept $j$ such that the semantic prototype $P_j\_upper$ is the most similar prototype to the BOW of this image region.
d. Compare the labels assigned to image regions from the test set with the ground truth labels. Report the results in the confusion matrix **mat_upper**.

**end**

**For $i$=1 to 10 Do:**

a. Randomly select 90% of all BOWs histograms, generated from the lower halves, for training and 10% for testing.
b. For each semantic concept $\bar{c}_j$, $j$=1,2,…, $N$, generate local semantic concept prototype $P_j\_lower$ by averaging all BOWs from the training set.
c. Use KNN classifier, using the Euclidean distance and K=1, to find similarities between the BOWs in the test set and the local semantic prototypes $P\_lower$. An image region at the lower halve, represented by a BOW histogram, is assigned to the semantic concept $j$ such that the semantic prototype $P_j\_lower$ is the most similar prototype to the BOW of this image region.
d. Compare the labels assigned to image regions from the test set with the ground truth labels. Report the results in the confusion matrix **mat_lower**.

**end**

**Output:** Confusion matrix **mat**, where **mat** = (**mat_upper**+**mat_lower**)

Three types of features are included: color histogram, color moments and DWT. These features are used separately to represent the visual content of image regions. Also, these features are linearly integrated with the BOW and IBOW histograms to study their influence on the discriminative power of BOW by including color and textural features.

The first set of experiments investigates image region annotation using local semantic prototypes and the KNN classifier. Algorithm 5 is used to generate local semantic prototypes from BOW histograms and to annotate new image regions with semantic concepts using the KNN classifier. The same algorithm can be applied to generate local prototypes for different features. The only thing that needs to be changed in the algorithm is to replace BOW with the features extracted from image regions. For example, replacing BOW with IBOW histograms will generate local semantic prototypes from IBOW and then the KNN classifier is used to annotate new image region, represented by an IBOW histogram, with one of the predefined semantic labels. Another example is to use color histogram to represent the visual content of image regions. In this work and similar to [5], three histograms are obtained from an image region represented in the HSV color space. The first histogram (36-bins) is obtained from the Hue component of the image region, the second histogram (32-bins) is obtained from the S component while the third histogram (16-bins) is obtained from the V component. These three histograms are concatenated to form a single HSV color histogram of 84-bins. Algorithm 5 is used again but BOW histograms are now replaced with the HSV color histograms from which local semantic prototypes are generated and used by the KNN classifier to annotate new image regions.

For multiple features, local semantic prototypes are first generated for each feature separately and then for each semantic concept the prototypes of the different feature types are aggregated to form a single local semantic prototype for the corresponding local semantic concepts. For example, the natural scene dataset used in this paper is composed of nine semantic concepts. Suppose that color histogram (84-D feature vector) and BOW (200-D feature vector) are the features to be concatenated. Firstly, Algorithm 5 is used to generate nine local semantic prototypes for the color histogram and the same algorithm is used to generate another nine local semantic prototypes for the BOW histograms. For each semantic concept, the local semantic prototype produced from the color histograms is concatenated with the local semantic prototype produced from the BOW histograms. This will generate nine local semantic prototypes each of length (84+200 = 284-D). The same approach can be applied for more than two feature types.

To this end, in the first experimental set, 14 experiments are conducted using the following features and their combinations for image region annotation: IBOW, UBOW, Color histogram (ColHist), color moments (Mom), DWT (Wav), IBOW+Mom, UBOW+Mom, IBOW+ColHist, UBOW+ColHist, IBOW+Wav, UBOW+Wav, IBOW+ColHist+Wav, UBOW+ColHist+Wav, and ColHist+Wav.

In the second set of experiments, SVM is used to annotate image regions. Here, local semantic prototypes are not used. For IBOW histograms, SVMs are trained on 90% of all image regions in the dataset, represented by their IBOW histograms. The trained SVMs are then used to assign labels to the remaining image regions in the dataset. For multiple features, each feature vector is first normalized to a unit length vector and then they are aggregated into a single feature vector. The SVMs are then used for training and testing. Similar to the first set of experiments, 14 experiments are conducted using the same features and their combination as aforementioned.

Figure 16 presents the performance of the first two sets of experiments. Bars in blue show the performance of image region annotation using local semantic prototypes with the KNN classifier. Bars in red show the performance of image region annotations using SVMS. It is worth to remind that IBOW and UBOW histograms are generated from visual vocabularies constructed at scene level and not concept level, the local from global approach. Thus, it is interesting to see their performances for image region annotation. It is obvious that SVMs outperforms the local semantic prototypes in all types of features and their combinations. Also, IBOW histograms outperform UBOW in both sets of experiments. Adding color and textural features to the UBOW and IBOW histograms improved the accuracy of image region annotation.

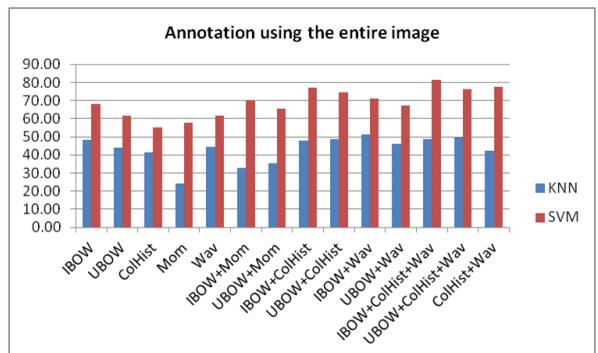

Figure 16: Accuracies of annotating images with the nine semantic concepts using KNN and SVM classifiers.

The performance of IBOW histograms has gained the best annotation accuracy when color histograms and DWT are added to them. Nevertheless, integrating low-level features such as color histogram with the DWT has also gained a good annotation performance compared with more complicated features such as IBOW and UBOW histograms. However, they have not got the best annotation results but they can be used to improve the performances of the IBOW and UBOW histograms. The annotation performance for each semantic concept resulted from the 14 experiments using both SVM and local semantic prototypes are shown in Table 2 and Table 3. Each row contains the accuracies of each semantic concept. These values are the diagonal elements of the confus ion matrix generated from each experiment. The last column from both tables is used to generate the results shown in Figure 16.

Table 2: (KNN and semantic prototypes) Accuracies of each experiment (row), where elements in each row are the diagonal elements of the confusion matrix resulted from each experiment. Accuracy is generated based on 10-folds CV.

|  | Sky | Water | Grass | Trunks | Foliage | Field | Rocks | Flowers | Sand | Acc. |
|---|---|---|---|---|---|---|---|---|---|---|
| IBOW | 88.54 | 34.50 | 9.93 | 37.35 | 45.56 | 25.93 | 32.56 | 44.22 | 26.42 | 48.41 |
| UBOW | 86.61 | 29.52 | 6.45 | 24.12 | 39.85 | 25.88 | 24.15 | 42.61 | 26.88 | 43.87 |
| ColHist | 58.91 | 49.69 | 61.00 | 24.25 | 26.79 | 43.46 | 22.10 | 49.15 | 48.19 | 41.19 |
| Mom | 36.24 | 29.12 | 3.94 | 35.75 | 15.22 | 32.95 | 10.68 | 42.80 | 43.21 | 24.20 |
| Wav | 84.68 | 29.51 | 22.44 | 38.65 | 47.06 | 16.05 | 16.45 | 48.22 | 29.34 | 44.47 |
| IBOW+Mom | 44.61 | 36.38 | 9.79 | 44.06 | 29.76 | 37.01 | 16.60 | 49.00 | 49.17 | 32.78 |
| UBOW+Mom | 48.76 | 35.69 | 13.90 | 42.89 | 35.43 | 35.72 | 18.29 | 49.19 | 50.26 | 35.50 |
| IBOW+ColHist | 65.29 | 54.93 | 63.49 | 36.62 | 36.47 | 46.08 | 29.55 | 54.32 | 51.75 | 47.92 |
| UBOW+ColHist | 66.63 | 54.48 | 59.12 | 38.40 | 40.53 | 46.20 | 29.31 | 55.30 | 51.69 | 48.92 |
| IBOW+Wav | 88.81 | 38.26 | 13.05 | 43.51 | 49.39 | 26.12 | 36.37 | 47.58 | 30.83 | 51.11 |
| UBOW+Wav | 86.47 | 33.85 | 8.02 | 30.34 | 43.23 | 26.48 | 26.50 | 44.75 | 30.83 | 46.05 |
| IBOW+ColHist+Wav | 66.32 | 55.44 | 64.37 | 37.48 | 37.59 | 46.37 | 30.61 | 54.86 | 52.84 | 48.84 |
| UBOW+ColHist+Wav | 67.62 | 55.00 | 60.38 | 39.08 | 41.59 | 46.63 | 29.98 | 55.78 | 52.95 | 49.77 |
| ColHist+Wav | 60.09 | 50.19 | 61.95 | 25.66 | 28.10 | 44.46 | 23.74 | 50.61 | 49.63 | 42.39 |

The last two sets of experiments are dedicated to study the influence of building visual vocabularies from local keypoints located at the upper and lower halves of images. It aims to improve the quality of visual words generated from clustering features of local keypoints at upper and lower halves of images.

Table 3: (SVM) Accuracies of each experiment (row), where elements in each row are the diagonal elements of the confusion matrix resulted from each experiment. Accuracy is generated based on 10-folds CV.

|  | Sky | Water | Grass | Trunks | Foliage | Field | Rocks | Flowers | Sand | Acc. |
|---|---|---|---|---|---|---|---|---|---|---|
| IBOW | 92.10 | 45.00 | 43.53 | 32.06 | 77.76 | 41.55 | 71.58 | 42.31 | 35.59 | 68.12 |
| UBOW | 91.59 | 45.30 | 24.12 | 37.35 | 76.98 | 13.04 | 67.40 | 24.40 | 3.50 | 61.65 |
| ColHist | 66.85 | 41.16 | 45.45 | 9.42 | 55.55 | 30.68 | 70.53 | 43.92 | 45.90 | 55.26 |
| Mom | 90.47 | 26.44 | 18.84 | 0.00 | 77.17 | 18.74 | 59.70 | 18.89 | 10.20 | 57.82 |
| Wav | 90.76 | 63.77 | 4.65 | 24.06 | 81.90 | 12.23 | 54.55 | 17.03 | 6.53 | 61.70 |
| IBOW+Mom | 94.70 | 47.44 | 42.62 | 37.78 | 81.22 | 45.37 | 72.86 | 27.38 | 42.29 | 70.20 |
| UBOW+Mom | 91.59 | 45.30 | 24.12 | 37.35 | 76.98 | 13.04 | 67.40 | 24.40 | 3.50 | 65.54 |
| IBOW+ColHist | 95.56 | 64.45 | 55.27 | 21.42 | 85.56 | 55.75 | 79.50 | 64.52 | 57.08 | 77.37 |
| UBOW+ColHist | 91.89 | 63.99 | 51.81 | 18.65 | 80.68 | 52.27 | 78.77 | 67.69 | 48.88 | 74.51 |
| IBOW+Wav | 93.89 | 64.80 | 38.28 | 32.86 | 80.64 | 40.19 | 71.56 | 50.37 | 20.57 | 71.12 |
| UBOW+Wav | 94.12 | 64.66 | 15.13 | 32.80 | 80.53 | 30.56 | 61.65 | 39.73 | 26.36 | 67.10 |
| IBOW+ColHist+Wav | 97.74 | 77.14 | 59.32 | 41.78 | 86.92 | 61.37 | 81.54 | 69.64 | 63.15 | 81.64 |
| UBOW+ColHist+Wav | 94.36 | 72.45 | 52.87 | 33.91 | 72.52 | 59.57 | 79.73 | 73.01 | 66.19 | 76.13 |
| ColHist+Wav | 95.93 | 68.57 | 55.75 | 38.77 | 75.00 | 67.93 | 82.14 | 71.79 | 56.05 | 77.61 |

Two visual vocabularies are generated from the upper halves of images; universal visual vocabulary and integrated visual vocabulary. Image regions located at the upper halves of images are represented by UBOW and IBOW histograms generated using the upper universal and integrated visual vocabularies, respectively. Another two visual vocabularies are generated from the lower halves of images. Image regions located at the lower halves of images are represented by UBOW and IBOW histograms generated using the lower universal and integrated visual vocabularies. To use multiple features with the UBOW and IBOW, the same low-level features used in the first two sets are employed here.

In the third set of experiments Algorithm 6 is used to generate local semantic prototypes for the upper halve and local semantic prototypes for the lower halve. Also, 14 experiments are conducted using this algorithm. Local semantic prototypes at the upper halve are used by KNN to annotate test image regions located at the upper halve of images. Results of annotations are compared with the ground truth and then reported in mat_upper confusion matrix. The same procedure is applied to the lower halve with results reported in another confusion matrix, mat_lower. Both matrices are added together to get a single confusion matrix. Results of annotating images at the upper and lower halves, using upper and lower local semantic prototypes and the KNN classifier, is shown in blue bars in Figure 17.

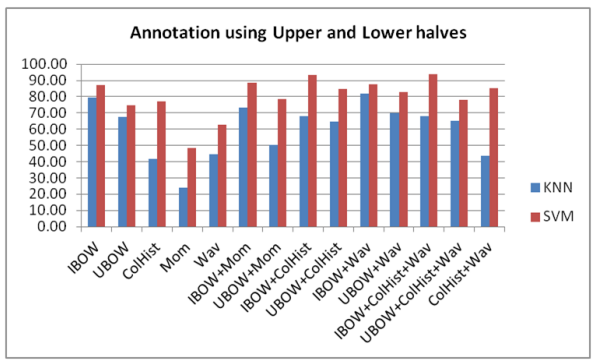

Figure 17: Accuracies of annotating images with the nine semantic concepts using KNN and SVM classifiers. BOWs and IBOWs are generated from image regions using visual vocabularies generated from the upper and lower halves of images.

It is obvious from the figure that generating local semantic prototypes from image halves improved the annotation accuracies in all experiments and for most features. The IBOW histograms are still the best in representing image regions. Adding textural features to IBOW has slightly improved the annotation and works better than adding color information. More details about the annotation accuracies of each of the experiments are shown in Table 4.

Table 4: (KNN and semantic prototypes at upper and lower halves) Accuracies of each experiment (row), where elements in each row are the diagonal elements of the confusion matrix (upper+lower) resulted from each experiment. Accuracy is generated based on 10-folds CV at each halves.

|  | Sky | Water | Grass | Trunks | Foliage | Field | Rocks | Flowers | Sand | Acc. |
|---|---|---|---|---|---|---|---|---|---|---|
| IBOW | 99.92 | 51.80 | 84.04 | 77.72 | 69.98 | 56.97 | 89.06 | 93.22 | 67.74 | 79.73 |
| UBOW | 98.09 | 29.77 | 68.28 | 55.88 | 55.39 | 36.75 | 81.41 | 67.40 | 47.45 | 67.56 |
| ColHist | 57.30 | 53.13 | 60.29 | 29.54 | 29.11 | 45.61 | 22.24 | 48.71 | 43.78 | 41.85 |
| Mom | 34.20 | 26.22 | 5.48 | 38.58 | 15.54 | 31.06 | 13.60 | 46.17 | 44.47 | 24.13 |
| Wav | 83.25 | 28.34 | 22.24 | 41.35 | 48.34 | 16.79 | 16.98 | 50.17 | 28.14 | 44.50 |
| IBOW+Mom | 60.13 | 60.77 | 87.70 | 80.12 | 74.63 | 61.77 | 91.60 | 93.51 | 75.99 | 73.16 |
| UBOW+Mom | 44.23 | 40.61 | 48.07 | 64.55 | 48.21 | 50.24 | 61.85 | 66.18 | 61.26 | 50.39 |
| IBOW+ColHist | 64.91 | 68.83 | 85.95 | 68.55 | 61.65 | 67.88 | 72.07 | 76.87 | 72.44 | 68.11 |
| UBOW+ColHist | 64.08 | 60.40 | 80.53 | 62.58 | 59.49 | 61.91 | 70.21 | 71.40 | 64.01 | 64.75 |
| IBOW+Wav | 86.71 | 63.57 | 85.90 | 83.75 | 82.23 | 62.63 | 90.19 | 93.85 | 75.42 | 81.70 |
| UBOW+Wav | 85.00 | 40.44 | 69.57 | 62.65 | 69.39 | 42.74 | 81.96 | 69.89 | 56.91 | 69.84 |
| IBOW+ColHist+Wav | 62.87 | 70.22 | 85.98 | 72.98 | 63.29 | 71.06 | 68.69 | 76.77 | 70.83 | 67.80 |
| UBOW+ColHist+Wav | 62.07 | 63.83 | 81.67 | 66.15 | 60.94 | 65.47 | 68.09 | 72.86 | 64.18 | 65.05 |
| ColHist+Wav | 55.08 | 56.56 | 61.55 | 37.66 | 34.13 | 51.03 | 22.88 | 46.80 | 50.72 | 43.75 |

The last set of experiments (14 experiments) is conducted on different features using SVM classifier. SVMs are trained and tested on image regions at the upper and lower halves of images. At the upper halve, image regions represented by the IBOW histograms or other features are used to train SVM which in turn used later to annotate test image regions at the upper halve of images. Results are reported in a confusion matrix mat_upper. The same approach is applied for image region at the lower halve of images. Results are reported in another confusion matrix mat_lower. Both matrices are add to each other to get final confusion matrix. The red bars in Figure 17 show annotations accuracies using different features. The best result is achieved using IBOW combined with ColHist and Wav features. It reports 94.02% annotation accuracy. More details are shown in Table 5.

Table 5: (SVM at upper and lower halves) Accuracies of each experiment (row), where elements in each row are the diagonal elements of the confusion matrix (upper+lower) resulted from each experiment. Accuracy is generated based on 10-folds CV at each halves.

|  | Sky | Water | Grass | Trunks | Foliage | Field | Rocks | Flowers | Sand | Acc. |
|---|---|---|---|---|---|---|---|---|---|---|
| IBOW | 90.44 | 82.96 | 99.46 | 89.42 | 76.53 | 68.55 | 99.55 | 99.90 | 82.35 | 87.17 |
| UBOW | 86.80 | 33.46 | 44.48 | 80.12 | 75.07 | 67.26 | 95.49 | 83.94 | 74.15 | 74.94 |
| ColHist | 94.36 | 65.94 | 58.15 | 10.77 | 84.85 | 59.15 | 75.05 | 71.45 | 66.65 | 76.88 |
| Mom | 85.28 | 0.40 | 1.68 | 11.63 | 47.38 | 4.92 | 77.30 | 20.35 | 24.30 | 48.45 |
| Wav | 91.72 | 66.02 | 9.31 | 24.25 | 81.88 | 16.95 | 52.35 | 18.01 | 12.84 | 62.61 |
| IBOW+Mom | 85.39 | 82.83 | 88.07 | 75.45 | 92.57 | 90.76 | 93.36 | 81.16 | 90.26 | 88.44 |
| UBOW+Mom | 86.17 | 41.82 | 50.56 | 57.29 | 82.23 | 67.86 | 99.52 | 99.66 | 80.80 | 78.65 |
| IBOW+ColHist | 95.86 | 85.58 | 99.49 | 87.75 | 96.21 | 89.57 | 94.24 | 85.31 | 95.99 | 93.61 |
| UBOW+ColHist | 86.80 | 86.25 | 93.01 | 73.97 | 76.53 | 72.04 | 94.61 | 83.94 | 79.89 | 84.57 |
| IBOW+Wav | 91.09 | 84.33 | 99.46 | 89.42 | 76.53 | 68.55 | 99.55 | 99.90 | 88.08 | 87.66 |
| UBOW+Wav | 86.15 | 85.56 | 90.15 | 73.97 | 75.07 | 67.26 | 94.16 | 79.06 | 74.15 | 83.09 |
| IBOW+ColHist+Wav | 96.51 | 85.58 | 99.49 | 87.75 | 96.21 | 89.57 | 94.24 | 92.63 | 95.99 | 94.02 |
| UBOW+ColHist+Wav | 88.08 | 41.23 | 59.12 | 58.09 | 75.84 | 63.49 | 99.72 | 99.66 | 86.82 | 78.04 |
| ColHist+Wav | 98.08 | 84.82 | 67.88 | 50.46 | 87.73 | 73.35 | 81.97 | 77.89 | 74.50 | 85.07 |

# 7. CONCLUSION

This paper has presented the problem of image annotation at image region level. The task was to assign labels to image regions generated from a regular grid. The paper has addressed using BOW model to represent image regions. A framework for image region annotation has been proposed. The relationship between the distributions of local semantic concepts and local keypoints located in image regions labelled with these semantic concepts are studied in detail. Also, this paper has investigated using visual vocabularies generated from natural scene classes to represent local semantic concepts, the local from global approach. Generating visual vocabularies from image halves were also investigated to generate BOW histograms. An extensive experimental work has been conducted using different features and classifiers to annotate image regions with semantic concepts. Our experimental results shows the plausibility of local from global approach for image region annotation as well as the discriminative power of using visual vocabularies from image halves. It showed an improved annotation results using IBOW combined with low-level features.

**ACKNOWLEDGEMENTS**

The author is grateful to the Applied Science Private University, Amman, Jordan, for the full financial support granted to this research. The author would like to thank Dr. Julia Vogel for providing us access to the natural scene image dataset and for valuable discussion.